\documentclass[conference]{IEEEtran}

\usepackage{tabularx}
\usepackage{marvosym}
\ifCLASSINFOpdf
  \usepackage[pdftex]{graphicx}
\else
  \usepackage[dvips]{graphicx}
\fi
\usepackage{tabularx}
\usepackage{graphicx}
\usepackage{subcaption}
\usepackage{amsmath}
\begin{document}
\title{Use neural networks to recognize students' handwritten letters and incorrect symbols}
\author{\IEEEauthorblockN{Jia Jun Zhu\textsuperscript{1} \;\;\; Binjie Hong\textsuperscript{1} \;\;\; Zichuan Yang\textsuperscript{1} \;\;\;
Jiacheng Song\textsuperscript{2,4}\;\;\; Jiwei Wang\textsuperscript{4} \;\;\; \\Tianhao Chen\textsuperscript{4}\;\;\; Shuilan Yang\textsuperscript{5} \;\;\;Zixun Lan\textsuperscript{3,4}\;\;\; Fei Ma\textsuperscript{3,\Letter}}

\IEEEauthorblockA{\textsuperscript{1}Department of Information and Computing Science, Xi'an Jiaotong-Liverpool University, Suzhou, China}
\IEEEauthorblockA{\textsuperscript{2}School of Engineering, Suzhou Centennial College, Suzhou, China}
\IEEEauthorblockA{\textsuperscript{3}Department of Applied Mathematics, Xi'an Jiaotong-Liverpool University, Suzhou, China}
\IEEEauthorblockA{\textsuperscript{4}GROW Inc., Suzhou, China}

\IEEEauthorblockA{\textsuperscript{5}TAL Education Group, Beijing, China}
\IEEEauthorblockA{\textsuperscript{\Letter}Corresponding author: Fei.Ma@xjtlu.edu.cn}}

% \author{\IEEEauthorblockN{JiaJun Zhu^1,a}
% \IEEEauthorblockA{School of Advanced Technology\\
% Xi’an Jiaotong-Liverpool University\\
% SuZhou China}
% \and
% \IEEEauthorblockN{Homer Simpson}
% \IEEEauthorblockA{Twentieth Century Fox\\
% Springfield, USA}
% \and
% \IEEEauthorblockN{James Kirk\\ and Montgomery Scott}
% \IEEEauthorblockA{Starfleet Academy\\
% San Francisco, California 96678-2391\\
% Telephone: (800) 555--1212\\


%Atlanta, Georgia 30332--0250}
%\IEEEauthorblockA{\IEEEauthorrefmark{2}Twentieth Century Fox, Springfield, USA}
%Telephone: (800) 555--1212, Fax: (888) 555--1212}


\maketitle

\begin{abstract}
%\boldmath
Correcting students' multiple-choice answers is a repetitive and mechanical task that can be considered an image multi-classification task. Assuming possible options are 'abcd' and the correct option is one of the four, some students may write incorrect symbols or options that do not exist. In this paper, five classifications were set up - four for possible correct options and one for other incorrect writing. This approach takes into account the possibility of non-standard writing options.
\end{abstract}


% \fi
\IEEEpeerreviewmaketitle

\section{Introduction}
Automated grading of multiple-choice questions is a repetitive and low-meaning task that deep learning-based computer vision has proven effective for. When dealing with deterministic multiple-choice questions, deep neural networks\cite{1} can be used for image recognition and treating the grading task as a Multiclass Classification task\cite{2}.
We proposes using deep learning to automate grading multiple-choice questions, treating it as a five-classification image recognition task. A significant flaw is that students may write random symbols leading to misjudgment, so an unknown category is added to the model. Two approaches are discussed for model selection: self-attention mechanism and MLPs\cite{3} or ResNet networks\cite{4}. The dataset is designed by selecting symbols without abcd features, using 26 letters data from the MNIST dataset\cite{5}. Details of the dataset division will be presented later in the article.

\section{Related Works}
Some algorithms, such as neural networks\cite{lan2021sub2}, decision trees, k-Nearest Neighbor, Naive Bayes, and Support Vector Machines, can naturally extend the binary classification technique to solve the multiclass classification problem. In other words, these algorithms can use the same approach for both binary and multiclass classification problems with some modifications.\cite{13}These background work of solving image classification tasks by predecessors can be roughly divided into traditional machine learning algorithms, which I will elaborate on in section II.A, and deep learning algorithms\cite{lan2023aednet} based on neural networks, which I will explain in section II.B.

\subsection{ traditional machine learning algorithms}

\subsubsection{Decision Trees}
Decision trees are a highly effective classification technique, with two well-known algorithms being Classification and Regression Trees (CART)\cite{14} and ID3/C4.5\cite{15}. These algorithms use available feature values to determine the best way to split the training data, producing a strong generalization. The split at each node is based on the feature that provides the most information gain. Each leaf node represents a class label, with new examples classified by following a path from the root node to a leaf node and testing certain features at each node. The leaf node reached determines the class label for the example. This algorithm is capable of handling binary or multiclass classification problems, and leaf nodes can refer to any of the K classes in question.
\subsubsection{k-Nearest Neighbors}
kNN\cite{16} is one of the earliest non-parametric classification algorithms. It classifies an unknown example by measuring its distance (using a distance measure like Eculidean) to every other training example. The algorithm identifies the k smallest distances and considers the output class label as the most represented class among these k classes. The value of k is usually determined using cross-validation or a validation set.
\subsubsection{Naive Bayes}
Naive Bayes \cite{17} is a successful classification algorithm that operates on the principle of Maximum A Posteriori (MAP). It works by assigning a class label c to an unknown example with features$\mathrm{x}=\left(x^{1}, \ldots, x^{n}\right)$ based on the maximum a posterior probability given the observed data. This probability is determined by the prior probabilities $\mathrm{P}\left(C_{1}\right), \ldots, \mathrm{P}\left(C_{k}\right)$ and the likelihood of the features given the class label. In summary, Naive Bayes chooses the class label with the highest probability given the observed data.
\subsubsection{Support Vector Machines}
Support Vector Machines (SVMs) are a classification algorithm known for their robustness and success \cite{18,19}. They work by maximizing the margin, which is the minimum distance between the separating hyperplane and the nearest example. SVMs typically only support binary classification, but researchers have proposed extensions \cite{20,21} to handle multiclass classification. These extensions add additional parameters and constraints to the optimization problem to separate the different classes. However, some formulations \cite{22} can result in a large optimization problem that may be impractical for a large number of classes. Other formulations, such as the one presented in \cite{23}, have a more efficient implementation.
\subsection{deep learning algorithms based on neural networks}
\subsubsection{convolutional neural networks}
 Convolutional Neural Networks (CNNs) are a popular type of deep learning architecture that mimics the visual perception mechanism found in living creatures . Hubel and Wiesel first discovered in 1959 that receptive fields in animal visual cortex cells are responsible for detecting light. Convolutional neural networks are highly effective for image recognition and classification tasks and four well-known CNNs from the past include AlexNet\cite{24}, VGGNet\cite{25}, GoogleNet\cite{26}, and ResNet\cite{27}. Over time, CNN architectures have become increasingly deeper, with ResNet - the ILSVRC 2015 champion - being about 20 times deeper than AlexNet and 8 times deeper than VGGNet. By increasing the depth of the network, it can better approximate the target function with increased nonlinearity and obtain improved feature representations.
 \subsubsection{vision transformer}
 Vision transoformer(ViT)\cite{28} is a model proposed by Google team in 2020, which applies Transformer\cite{29}to image calssification.Although not the first paper to apply Transformer to visual tasks, ViT has become a milestone work for the application of Transformer in the field of computer vision due to its ‘simple’ model, good performance, and strong scalability(the larger the model\cite{lan2021sub}, the better the performance), and has sparked subsequent related research. The most prominent feature of ViT is that when there is enough data for pre-training, its performance exceeds that of CNN, breaking the restriction of Transformer’s lack of inductive bias and achieving good transfer effects\cite{ma2021global} in downstream tasks. However,when the training dataset is not large enough, the performance of ViT is usually worse than that of ResNets\cite{27} of the same size because Transformer lacks inductive bias compared to CNN,namely a kind of prior knowledge or assumption made in advance. CNN has two kinds of inductive bias: one is locality(two-dimensional neighborhood structure), that is, adjacent regions on the image have similar features. The other is translation equivariance, f(g(x))=g(f(x)), where g represents convolution operation and f represents translation operation.When CNN has these two kinds of inductive bias, it has a lot of prior information and can learn a relatively good model with relatively little data. 
 \section{Methodology}
 \subsection{Models}
 As stated in the introduction, we consider the task of identifying and grading student multiple-choice answers as an image multi-classification task. To avoid misidentifying symbols outside of the given options as correct answers, we have added an additional category to our model. Specifically, assuming that there are four correct options to be graded in multiple-choice questions, this task will be treated as a five-classification task, with the extra category being for identifying non-conforming writing options in student responses.For the purpose of conducting more efficient comparative experiments in this task, we have simplified the actual problem. Specifically, in this paper, all multiple-choice questions in the applications considered are treated as single-choice questions with four options. As a result, our task has been simplified to an image recognition five-classification task.We ultimately chose to utilize ResNet as our primary model for completing our task. It is commonly believed that the ability of a network to extract features is enhanced to a certain extent as the number of layers increases. ResNet's Shortcut Connections residual connection method can effectively help us make the network deeper. The fundamental principle of residual connections is to add the input x to the output F(x) after residual convolution, resulting in a final output that contains all the information from the input and output ends. This can only be done if the dimensions of the output end and input end are consistent, as defined:
\begin{figure}[htbp]
  \centering
  \includegraphics[width=0.4\textwidth]{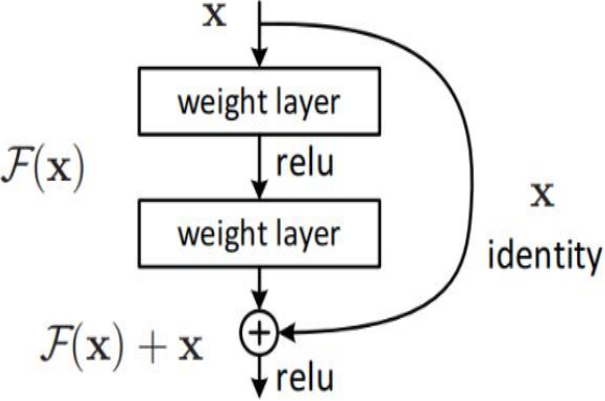}
  \caption{a building block for Resnet\cite{27}}
  \label{2}
\end{figure}
In Eqn. (1), x and y represent the input and output vectors of the layers considered. The function $\mathrm{F}\left(\mathrm{x},\left\{\mathrm{W}_{\mathrm{s}}\right\}\right)$ represents the residual mapping that needs to be learned. For the example in Fig. 1 with 
two layers, $\mathrm{F}=\mathrm{W}_{2} \sigma\left(\mathrm{W}_{1} \mathrm{x}\right)$ where $\sigma$ epresents the ReLU\cite{30}nonlinearity and biases are omitted for simplification. The operation of  F + x is performed by a shortcut connection and element-wise addition. We adopt the second nonlinearity after the addition,such as $\sigma(\mathrm{y})$in Fig. 1\cite{27}. To summarize, the shortcut connections in ResNet allow for deeper networks without introducing extra parameters or computation complexity. Linear projection can be used to match the dimensions of x and F if needed.
In Eqn. (1), it is necessary for the dimensions of x and F to be identical. In the event that this is not already the case, we can utilize a linear projection technique to manipulate the results and ensure that the dimensions are matched appropriately:
\begin{equation}
    \mathrm{y}=\mathrm{F}\left(\mathrm{x},\left\{\mathrm{W}_{\mathrm{s}}\right\}\right)+\mathrm{W}_{\mathrm{i}} \mathrm{x}
\end{equation}
ResNet can be built with deeper layers and better feature extraction using shortcut connections. The common ResNet models have 18, 50, 101, and 152 layers. After conducting multiple ablation experiments for our image classification task, it has been proven that ResNet50 is the most suitable model for our task. The trained ResNet50 achieved the highest accuracy of 98.6\% on our validation set. This accuracy did not improve with deeper ResNet101 and ResNet152 models. Therefore, we believe that ResNet50 is the most economical choice for our image recognition five-classification task. Further details of the ablation experiments will be elaborated in section 4. The approximate model architecture of ResNet50 that we have chosen is presented below as the Fig. 2.
\begin{figure}[htbp]
  \centering
  \includegraphics[width=0.4\textwidth]{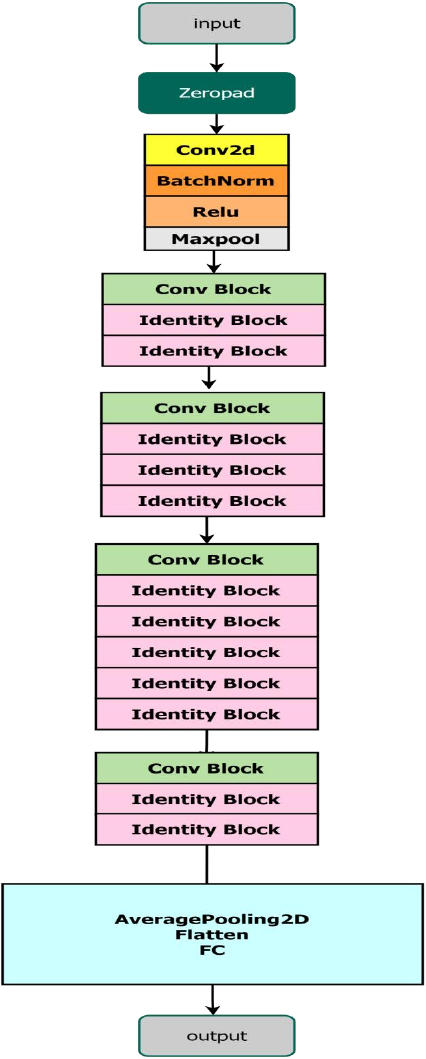}
  \caption{Example network architectures for Resnet50}
  \label{2}
\end{figure}
  As we can see in Fig. 2. ResNet50 is composed of 49 convolutional layers and one fully connected layer. The network structure can be divided into seven parts, with the first part performing convolution, regularization, activation function, and maximum pooling calculations on the input. The second to fifth parts contain residual blocks, with the green blocks changing the dimension but not the size of the residual block. Each residual block has three convolutional layers, resulting in a total of 49 convolutional layers and one fully connected layer. The input to the network is 224*224*3, and after the convolution calculations, the output is 7*7*2048. The pooling layer converts the output into a feature vector, which is then used by the classifier to calculate and output the class probabilities.
 \subsection{3.2Dataset}
 Our dataset was selected from the English letter dataset in the public dataset MNIST\cite{31}, as mentioned in the Introduction. Our task is to recognize the answers to multiple-choice questions written by students by hand. Given that most of the answers to multiple-choice questions are primarily in English letters, the handwritten English letters in MNIST are very suitable for our requirements. As previously mentioned, we simplify our grading application scenario to four-choice questions (questions with more options can easily be extended with our method). However, we recognize that in addition to setting the basic four possible answers of ABCD for classification on our model, we must also prevent students' incorrectly written and non-standard symbols from being randomly classified by the machine as options in ABCD, leading to grading errors. Therefore, we added a fifth class on top of the four-class classification, named "unknown," indicating that answers written by students recognized as belonging to this class are not any of the ABCD options. Regarding this innovation, we also made a more reasonable division of the MNIST dataset. We selected the images of the four handwritten English letters ABCD to construct the first four categories of our datasets, and randomly and evenly selected another 22 English letters to construct a fifth category, which is the same size as each of the ABCD categories, called the "unknown" category, used to recognize students' non-standard writing. The underlying reason for such division is that in essence, constructing the fifth category to recognize students' non-standard writing only requires selecting a large number of diverse data images that are different from ABCD, and the datasets of the fifth category composed of the remaining 22 letters\cite{lan2022more} meets this criterion. The shapes of the 22 letters are diverse and different from the ABCD letters, and the model can distinguish and recognize symbols written by students that are not the prescribed ABCD four letters after learning a large amount of letters that do not belong to ABCD.The details of the datasets partitioning are illustrated in the TABLE 1.
\begin{table}[!ht]
    \centering
    \caption{Our dataset Selected from MNIST}
    
    \begin{tabular}{|p{0.1\textwidth}|p{0.04\textwidth}|p{0.04\textwidth}|p{0.04\textwidth}|p{0.04\textwidth}|p{0.05\textwidth}|}
    \hline
        CLASS & Class A & Class B & Class C & Class D & Class unknow  \\ \hline
        Train datasets & 4800 & 4800 & 4800 & 4800 & 4800  \\ \hline
        Val datasets & 800 & 800 & 800 & 800 & 800 \\ \hline
    \end{tabular}
\end{table}
\subsection{Training strategy}
\subsubsection{Transfer Learning}
In order to train our task model better and faster, we will introduce pre-trained networks such as ResNet50, VGG, and AlexNet provided in torchvision.models during model training. We will use transfer learning to fix the weights of all layers in the front of the pre-trained model and only train the last layer. We will also fully train two networks for comparison. More details will be provided in Section. Experiments.

\subsubsection{Cyclic Cosine Annealing Learning Rate Schedule}
To enhance our network training, we will apply a learning rate decay strategy called Cyclic Cosine Annealing Learning Rate Schedule\cite{33}. This strategy enables our learning rate to periodically fluctuate during the training process, resulting in better training performance,which defined as below:

\begin{equation}
    \eta_{t}=\eta_{\text {min }}^{i}+\frac{1}{2}\left(\eta_{\text {max }}^{i}-\eta_{\text {min }}^{i}\right)\left(1+\cos \left(\frac{\text { Tur }^{\text {cur }}}{T_{i}} \pi\right)\right)
\end{equation}
\subsection{loss function}
We choose the Cross Cross Entropy Loss Function as our loss function. It is a commonly used loss function in machine learning for classification tasks. It is used to measure the difference between the predicted probability distribution and the true probability distribution of a classification problem. The function is defined as the negative sum of the true class probability multiplied by the logarithm of the predicted class probability. The formula for Cross Entropy Loss Function is as follows:
\begin{equation}
    H(p, q)=-\sum_{x} P(x) \log (q(x))
\end{equation}
where P(x) is the true probability distribution and$q(x)$is the predicted probability distribution. The function penalizes the model more when it predicts a low probability for the true class and assigns high probabilities to other classes.
\section{Experiments}
In this section, we study the performance of our designed task on our selected datasets and 
demonstrate its effectiveness and efficiency compared with different models.
\subsection{Preprocessing}
As described in Section 3.2, our dataset is derived from the MNIST dataset and has been divided into training and validation sets, each containing five classes. The specifications of each data point in the training and validation sets are identical, with 28x28 pixel grayscale images. Before inputting the data, we preprocess the image data by resizing the images to 64x64 pixels, believing that larger images will make it easier for the model to extract information and improve the classification accuracy of our task.
\subsection{Parameter setting}
During the experiments, we standardized some basic parameters to better compare the effectiveness of different models in accomplishing our designed image five-classification task. The parameters we set include a batch size of 128 and 20 epochs. Regarding the learning rate, we adopted the Cyclic Cosine Annealing Learning Rate decay strategy, as mentioned in Section 3.3. The initial learning rate was set at 0.01 and the minimum learning rate was designed to be 0.00001. Throughout the training process, the learning rate oscillates periodically between 0.01 and 0.00001, with a half oscillation cycle occurring every six epochs. This approach allows for a more comprehensive evaluation of the models' performance in our classification task.
\section{Results}
\subsection{Comparison Experiment}
perform our task, with the sole criterion for evaluating the excellence of a model\cite{lan2023rcsearcher} being the accuracy on the validation set.During the training process, we will record the best accuracy achieved across all epochs to represent the capabilities of each model. We will compare the training of five models suitable for image classification tasks, including pretrained 18-layer and 50-layer ResNets, AlexNet, 19-layer VGG, and ViT a kind of Vision Transformer. For each model, we have tested the effectiveness of transfer learning, adopting the following training strategies: 1. Apply transfer learning by training only the last output layer of each model while using the pretrained parameters for the remaining layers of the network. 2. Load the parameters from the first training, unlock all layers of the model, and train all parameters completely. 3. Do not use transfer learning, and train the entire model from scratch.
% \begin{table}[!ht]
%     \centering
%     \caption{compare different models}
%     \begin{tabular}{|l|l|l|l|}
%     \hline
%         Training strategy & transfer learning & Retrain the whole  & Train from scratch  \\ \hline
%         Resnet18 & 80.60\% & 98.25\% & 98.48\%  \\ \hline
%         Resnet50 & 84.25\% & 99.40\% & 98.88\%  \\ \hline
%         Vgg19 & 83.00\% & 98.33\% & 97.9\%  \\ \hline
%         ViT & 63.20\% & 77.42\% & 70.12\%  \\ \hline
%         Alexnet & 84.70\% & 88.41\% & 80.12\% \\ \hline
%     \end{tabular}
% \end{table}
\begin{table}[!ht]
    \centering
    \caption{compare different models}
    \scalebox{0.85}{
    \begin{tabular}{|l|l|l|l|}
    \hline
        Training strategy & transfer learning & Retrain the whole  & Train from scratch  \\ \hline
        Resnet18 & 80.60\% & 98.25\% & 98.48\%  \\ \hline
        Resnet50 & 84.25\% & 99.40\% & 98.88\%  \\ \hline
        Vgg19 & 83.00\% & 98.33\% & 97.9\%  \\ \hline
        ViT & 63.20\% & 77.42\% & 70.12\%  \\ \hline
        Alexnet & 84.70\% & 88.41\% & 80.12\% \\ \hline
    \end{tabular}%
    }
\end{table}
The content of the Table 2. represents the accuracy on the validation set after training different models. Each data point is rounded to two decimal places. It can be observed that, in general, training with transfer learning results in higher accuracy compared to training the entire model from scratch. Among the different models, except for Vision Transformer, the best training strategies yield very similar accuracy rates. However, ResNet50 achieves slightly better results compared to the other models.As shown in below, we present more detailed training data for the best-performing model, Resnet50.
\begin{figure}[htbp]
  \centering
  \includegraphics[width=0.4\textwidth]{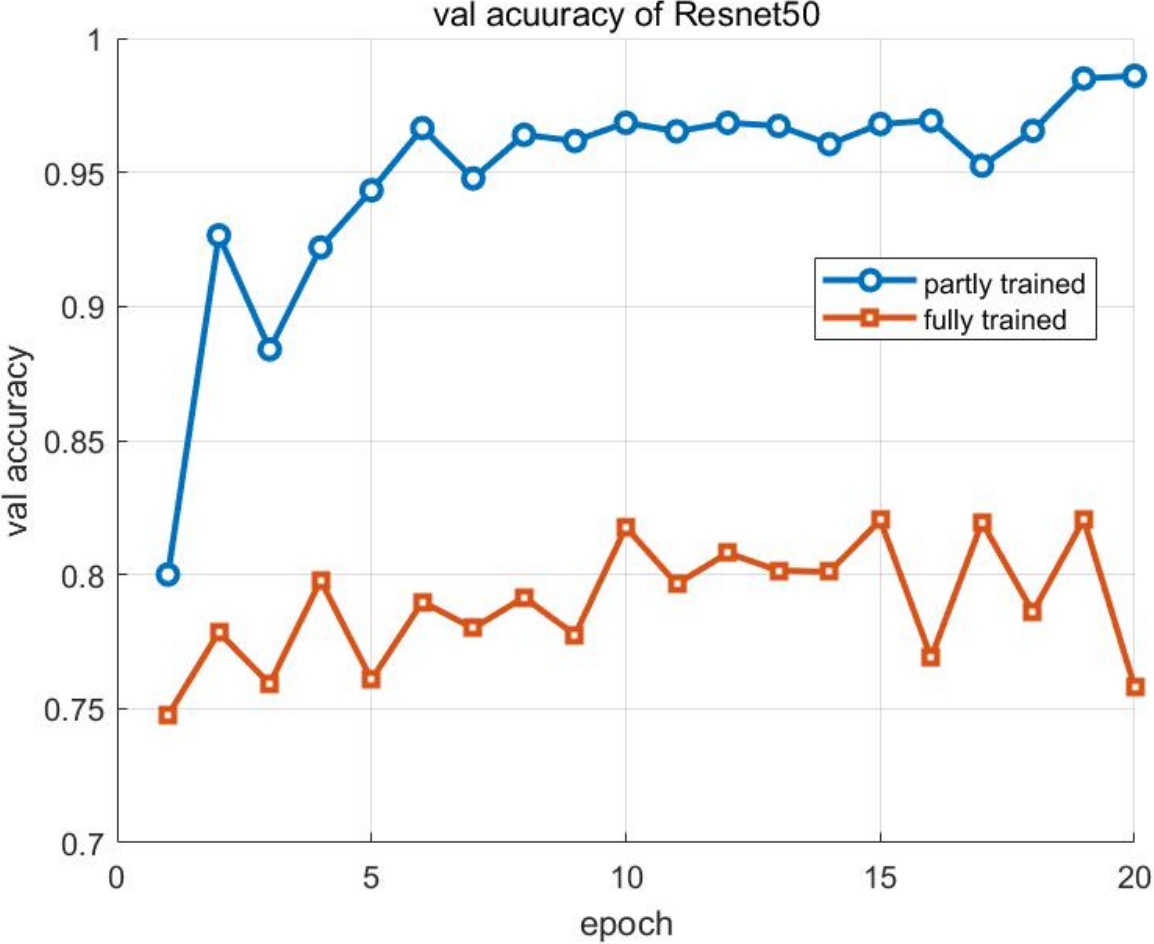}
  \caption{the val accuracy of resnet50}
  \label{2}
\end{figure}
The above figure provides a detailed representation of the training process of Resnet50 over 20 epochs. Our sole criterion for evaluating the model is the accuracy on the validation set, which is the percentage of correctly classified samples in the validation set. The figure displays two curves: the orange curve represents the validation accuracy of the Resnet50 model trained only on the linear output layer, while the blue curve represents the validation accuracy of the model obtained through transfer learning by fully training it after inheriting the weights of the last linear layer. It is evident that the fully trained model yields significantly higher accuracy compared to the partially trained model.
\subsection{Results Analysis}
Using the best-performing ResNet50, we can achieve an accuracy of up to 98.6\% on the validation set. In fact, after analyzing the data, we believe that our combined training strategy has even higher accuracy potential. Upon analyzing the misclassified images, we found that one obvious reason is the presence of some difficult-to-identify characters in the MNIST dataset, as shown in the following image. In reality, the model obtained using our method has an accuracy of at least 99\% when grading real-world multiple-choice questions, and it can accurately identify students' incorrectly written characters without misjudging them.

\begin{figure}
  \centering
  \begin{subfigure}[b]{0.2\textwidth}
    \includegraphics[width=\textwidth]{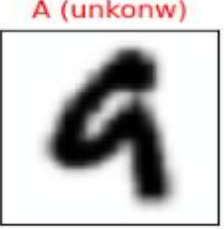}
    \caption{}
    \label{fig:sub1}
  \end{subfigure}
  \hfill
  \begin{subfigure}[b]{0.2\textwidth}
    \includegraphics[width=\textwidth]{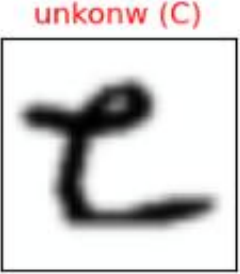}
    \caption{}
    \label{fig:sub2}
  \end{subfigure}
  \hfill
  \begin{subfigure}[b]{0.2\textwidth}
    \includegraphics[width=\textwidth]{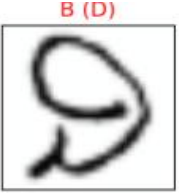}
    \caption{}
    \label{fig:sub3}
  \end{subfigure}
  \hfill
  \caption{the Misclassified Datas}
  \label{fig:group}
\end{figure}
The samples shown in Fig.4 are all misclassified. The model's predicted answers are marked outside the brackets, while the true labels are inside the brackets. These misclassified samples are only a portion of all incorrect data. Upon analyzing them, we found that even with the human eye, it is difficult to accurately determine their true labels. A part of our error set consists of such cases, so we believe that the actual accuracy of our task under the combined strategy is even higher.
\section{Conclusion}
In this paper, we mainly discussed how to better use AI to replace manual recognition of students’ written answers and give judgments. We regard the landing application of this AI in the field of education as an image multi-classification task. Due to the possibility of students writing non-standard symbols incorrectly, we added an “unknown” class after each correct answer corresponding category to identify non-standard writing options. In training such a model, we compared many famous models horizontally and used many training strategies and finally decided that ResNet50 is our local optimal solution. In the future, we will continue to explore better ways to expand the application of AI intelligent correction, and think and deliberate on more application scenarios.


\begin{thebibliography}{10}

\bibitem{1}
V.~Deved{\v{z}}i{\'c}, ``Web intelligence and artificial intelligence in education,'' {\em Journal of Educational Technology \& Society}, vol.~7, no.~4, pp.~29--39, 2004.

\bibitem{2}
M.~J. Timms, ``Letting artificial intelligence in education out of the box: educational cobots and smart classrooms,'' {\em International Journal of Artificial Intelligence in Education}, vol.~26, pp.~701--712, 2016.

\bibitem{3}
L.~Chen, P.~Chen, and Z.~Lin, ``Artificial intelligence in education: A review,'' {\em Ieee Access}, vol.~8, pp.~75264--75278, 2020.

\bibitem{4}
S.~Bonthu, S.~Rama~Sree, and M.~Krishna~Prasad, ``Automated short answer grading using deep learning: A survey,'' in {\em Machine Learning and Knowledge Extraction: 5th IFIP TC 5, TC 12, WG 8.4, WG 8.9, WG 12.9 International Cross-Domain Conference, CD-MAKE 2021, Virtual Event, August 17--20, 2021, Proceedings 5}, pp.~61--78, Springer, 2021.

\bibitem{5}
A.~Voulodimos, N.~Doulamis, A.~Doulamis, E.~Protopapadakis, {\em et~al.}, ``Deep learning for computer vision: A brief review,'' {\em Computational intelligence and neuroscience}, vol.~2018, 2018.

\bibitem{6}
B.~M. Alsafy, Z.~M. Aydam, and W.~K. Mutlag, ``Multiclass classification methods: a review,'' {\em International Journal of Advanced Engeineering Technology And Innovative Science}, vol.~5, no.~3, pp.~1--10, 2014.

\bibitem{7}
J.~Schmidhuber, ``Deep learning in neural networks: An overview,'' {\em Neural networks}, vol.~61, pp.~85--117, 2015.

\bibitem{8}
K.~He, X.~Zhang, S.~Ren, and J.~Sun, ``Deep residual learning for image recognition,'' in {\em Proceedings of the IEEE conference on computer vision and pattern recognition}, pp.~770--778, 2016.

\bibitem{9}
M.~D. Zeiler and R.~Fergus, ``Visualizing and understanding convolutional networks,'' in {\em Computer Vision--ECCV 2014: 13th European Conference, Zurich, Switzerland, September 6-12, 2014, Proceedings, Part I 13}, pp.~818--833, Springer, 2014.

\bibitem{10}
A.~Dosovitskiy, L.~Beyer, A.~Kolesnikov, D.~Weissenborn, X.~Zhai, T.~Unterthiner, M.~Dehghani, M.~Minderer, G.~Heigold, S.~Gelly, {\em et~al.}, ``An image is worth 16x16 words: Transformers for image recognition at scale,'' {\em arXiv preprint arXiv:2010.11929}, 2020.

\bibitem{11}
K.~He, X.~Zhang, S.~Ren, and J.~Sun, ``Deep residual learning for image recognition,'' in {\em Proceedings of the IEEE conference on computer vision and pattern recognition}, pp.~770--778, 2016.

\bibitem{12}
Y.~LeCun, ``The mnist database of handwritten digits,'' {\em http://yann. lecun. com/exdb/mnist/}, 1998.

\bibitem{13}
M.~Aly, ``Survey on multiclass classification methods,'' {\em Neural Netw}, vol.~19, no.~1-9, p.~2, 2005.

\bibitem{14}
A.~Trendowicz, R.~Jeffery, A.~Trendowicz, and R.~Jeffery, ``Classification and regression trees,'' {\em Software Project Effort Estimation: Foundations and Best Practice Guidelines for Success}, pp.~295--304, 2014.

\bibitem{15}
J.~R. Quinlan, ``Program for machine learning,'' {\em C4. 5}, 1993.

\bibitem{16}
S.~D. Bay, ``Combining nearest neighbor classifiers through multiple feature subsets.,'' in {\em ICML}, vol.~98, pp.~37--45, Citeseer, 1998.

\bibitem{17}
I.~Rish {\em et~al.}, ``An empirical study of the naive bayes classifier,'' in {\em IJCAI 2001 workshop on empirical methods in artificial intelligence}, vol.~3, pp.~41--46, 2001.

\bibitem{18}
I.~Rish {\em et~al.}, ``An empirical study of the naive bayes classifier,'' in {\em IJCAI 2001 workshop on empirical methods in artificial intelligence}, vol.~3, pp.~41--46, 2001.

\bibitem{19}
C.~J. Burges, ``A tutorial on support vector machines for pattern recognition,'' {\em Data mining and knowledge discovery}, vol.~2, no.~2, pp.~121--167, 1998.

\bibitem{20}
J.~Weston and C.~Watkins, ``Multi-class support vector machines,'' tech. rep., Citeseer, 1998.

\bibitem{21}
E.~J. Bredensteiner and K.~P. Bennett, ``Multicategory classification by support vector machines,'' {\em Computational Optimization: A Tribute to Olvi Mangasarian Volume I}, pp.~53--79, 1999.

\bibitem{22}
J.~Weston, C.~Watkins, {\em et~al.}, ``Support vector machines for multi-class pattern recognition.,'' in {\em Esann}, vol.~99, pp.~219--224, 1999.

\bibitem{23}
K.~Crammer and Y.~Singer, ``On the algorithmic implementation of multiclass kernel-based vector machines,'' {\em Journal of machine learning research}, vol.~2, no.~Dec, pp.~265--292, 2001.

\bibitem{24}
O.~Russakovsky, J.~Deng, H.~Su, J.~Krause, S.~Satheesh, S.~Ma, Z.~Huang, A.~Karpathy, A.~Khosla, M.~Bernstein, {\em et~al.}, ``Imagenet large scale visual recognition challenge,'' {\em International journal of computer vision}, vol.~115, pp.~211--252, 2015.

\bibitem{25}
K.~Simonyan and A.~Zisserman, ``Very deep convolutional networks for large-scale image recognition,'' {\em arXiv preprint arXiv:1409.1556}, 2014.

\bibitem{26}
C.~Szegedy, W.~Liu, Y.~Jia, P.~Sermanet, S.~Reed, D.~Anguelov, D.~Erhan, V.~Vanhoucke, and A.~Rabinovich, ``Going deeper with convolutions,'' in {\em Proceedings of the IEEE conference on computer vision and pattern recognition}, pp.~1--9, 2015.

\bibitem{27}
C.~Szegedy, W.~Liu, Y.~Jia, P.~Sermanet, S.~Reed, D.~Anguelov, D.~Erhan, V.~Vanhoucke, and A.~Rabinovich, ``Going deeper with convolutions,'' in {\em Proceedings of the IEEE conference on computer vision and pattern recognition}, pp.~1--9, 2015.

\bibitem{28}
A.~Dosovitskiy, L.~Beyer, A.~Kolesnikov, D.~Weissenborn, X.~Zhai, T.~Unterthiner, M.~Dehghani, M.~Minderer, G.~Heigold, S.~Gelly, {\em et~al.}, ``An image is worth 16x16 words: Transformers for image recognition at scale,'' {\em arXiv preprint arXiv:2010.11929}, 2020.

\bibitem{29}
A.~Vaswani, N.~Shazeer, N.~Parmar, J.~Uszkoreit, L.~Jones, A.~N. Gomez, {\L}.~Kaiser, and I.~Polosukhin, ``Attention is all you need,'' {\em Advances in neural information processing systems}, vol.~30, 2017.

\bibitem{30}
V.~Nair and G.~E. Hinton, ``Rectified linear units improve restricted boltzmann machines,'' in {\em Proceedings of the 27th international conference on machine learning (ICML-10)}, pp.~807--814, 2010.

\bibitem{31}
L.~Deng, ``The mnist database of handwritten digit images for machine learning research [best of the web],'' {\em IEEE signal processing magazine}, vol.~29, no.~6, pp.~141--142, 2012.

\bibitem{32}
S.~J. Pan and Q.~Yang, ``A survey on transfer learning,'' {\em IEEE Transactions on knowledge and data engineering}, vol.~22, no.~10, pp.~1345--1359, 2009.

\bibitem{33}
I.~Loshchilov and F.~Hutter, ``Sgdr: Stochastic gradient descent with warm restarts,'' {\em arXiv preprint arXiv:1608.03983}, 2016.

\bibitem{lan2021sub}
Z.~Lan, L.~Yu, L.~Yuan, Z.~Wu, Q.~Niu, and F.~Ma, ``Sub-gmn: The subgraph matching network model,'' {\em arXiv preprint arXiv:2104.00186}, 2021.

\bibitem{lan2023aednet}
Z.~Lan, Y.~Ma, L.~Yu, L.~Yuan, and F.~Ma, ``Aednet: Adaptive edge-deleting network for subgraph matching,'' {\em Pattern Recognition}, vol.~133, p.~109033, 2023.

\bibitem{ma2021global}
Y.~Ma, Z.~Lan, L.~Zong, and K.~Huang, ``Global-aware beam search for neural abstractive summarization,'' {\em Advances in Neural Information Processing Systems}, vol.~34, pp.~16545--16557, 2021.

\bibitem{lan2022more}
Z.~Lan, B.~Hong, Y.~Ma, and F.~Ma, ``More interpretable graph similarity computation via maximum common subgraph inference,'' {\em arXiv preprint arXiv:2208.04580}, 2022.

\bibitem{lan2023rcsearcher}
Z.~Lan, Z.~Zeng, B.~Hong, Z.~Liu, and F.~Ma, ``Rcsearcher: Reaction center identification in retrosynthesis via deep q-learning,'' {\em arXiv preprint arXiv:2301.12071}, 2023.

\bibitem{lan2021sub2}
Z.~Lan, L.~Yu, L.~Yuan, Z.~Wu, Q.~Niu, and F.~Ma, ``Sub-gmn: The neural subgraph matching network model,'' {\em arXiv preprint arXiv:2104.00186}, 2021.

\end{thebibliography}
\end{document}